\documentclass[letterpaper, 10 pt, conference]{ieeeconf} 
\IEEEoverridecommandlockouts                      
\usepackage{cite}
\usepackage{amsmath,amssymb,amsfonts}
\usepackage{algorithmic}
\usepackage{graphicx}
\usepackage{textcomp}
\usepackage{xcolor}
\usepackage{balance}
\usepackage{subcaption}
\usepackage{bm}
\usepackage{booktabs}

\title{\LARGE \bf
Supervision and Source Domain Impact on Representation Learning: \\A Histopathology Case Study
}

\author{Milad Sikaroudi$^{1}$, Amir Safarpoor$^{1}$, Benyamin Ghojogh$^{2}$, Sobhan Shafiei$^{1}$,\\
Mark Crowley$^{2}$ and  H.R. Tizhoosh$^{1,*}$, {\it Senior Member, IEEE}% <-this % stops a space
\thanks{$^{1}$M. Sikaroudi, A. Safarpoor, S. Shafiei and H.R. Tizhoosh are  members of the  Laboratory  for  Knowledge  Inference  in  Medical  Image  Analysis (Kimia Lab), University of Waterloo, Waterloo, ON, Canada, 
{(e-mail: \{msikaroudi, asafarpo, s7shafie, tizhoosh\}@uwaterloo.ca}).}%
\thanks{$^{2}$B. Ghojogh and M. Crowley are with the Department of Electrical and Computer Engineering, University of Waterloo, Waterloo, ON, Canada, {(e-mail: \{bghojogh, mcrowley\}@uwaterloo.ca}).}%
\thanks{The first three authors contributed equally to this work.}
\thanks{$^{*}$Corresponding author: H.R. Tizhoosh.}        
}
\usepackage{eso-pic}

\begin{document}
\AddToShipoutPictureBG*{%
  \AtPageUpperLeft{%
    \setlength\unitlength{1in}%
    \hspace*{\dimexpr0.5\paperwidth\relax}
    \makebox(0,-0.75)[c]{\small Accepted for presentation at the 42nd Annual International Conference of the IEEE Engineering in Medicine and Biology Society (EMBC'20)}
    }}

\bstctlcite{IEEEexample:BSTcontrol}

\maketitle
\thispagestyle{empty}
\pagestyle{empty}

%%%%%%%%%%%%%%%%%%%%%%%%%%%%%%%%%%%%%%%%%%%%%%%%%%%%%%%%%%%%%%%%%%%%
\begin{abstract}
As many algorithms depend on a suitable representation of data, learning unique features is considered a crucial task. Although supervised techniques using deep neural networks have boosted the performance of representation learning, the need for a large sets of labeled data limits the application of such methods. As an example, high-quality delineations of regions of interest in the field of pathology is a tedious and  time-consuming task due to the large image dimensions. In this work, we explored the performance of a deep neural network and triplet loss in the area of representation learning. We investigated the notion of similarity and dissimilarity in pathology whole-slide images and compared different setups from unsupervised and semi-supervised to supervised learning in our experiments. Additionally, different approaches were tested, applying few-shot learning on two publicly available pathology image datasets. We achieved high accuracy and generalization when the learned representations were applied to two different pathology datasets.
\end{abstract}

\section{Introduction}
With the advent of digital whole-slide image (WSI) scanners and digital pathology, a vast range of computer-vision algorithms using machine learning have been developed to process histopathology images. These technologies offer opportunities for better quantitative modeling of disease appearance and hence possibly improved prediction of disease severity and aggressiveness, and patient outcome. More specifically, machine-learning applications in digital pathology ranges from computer-aided diagnosis based on classification, detection, segmentation, and content-based image retrieval. But domain-specific limitations such as large image dimensions and insufficient amount of annotated data restrict the applicability of such approaches \cite{komura2018machine}.

Current deep learning algorithms may reach or surpass human-level accuracy when a large amount of data is available. However, deep networks suffer from poor sample efficiency in stark contrast to human perception that can learn object categories after seeing just a few pictures, in some cases even one instance. Few-shot learning as the ability to learn from a few labeled samples aims to address this issue. More specifically, the knowledge is extracted from other similar problems since there is not enough data. As a result, many methods characterize few-shot learning as a meta-learning problem.

Approaches to meta-learning can fall into three major sub-categories by whether they rely on prior knowledge about similarity, learning, or data \cite{vanschoren2018meta}. We aimed to utilize prior knowledge about the similarity to learn robust embeddings by investigating the notion of pairwise similarity between samples. To this end, we propose a novel framework to impose abstract domain knowledge in histopathology as prior knowledge to train a triplet deep neural network.

\begin{figure*}[ht!]
	\centering
	\begin{subfigure}{0.6\textwidth}
		\includegraphics[width=\textwidth]{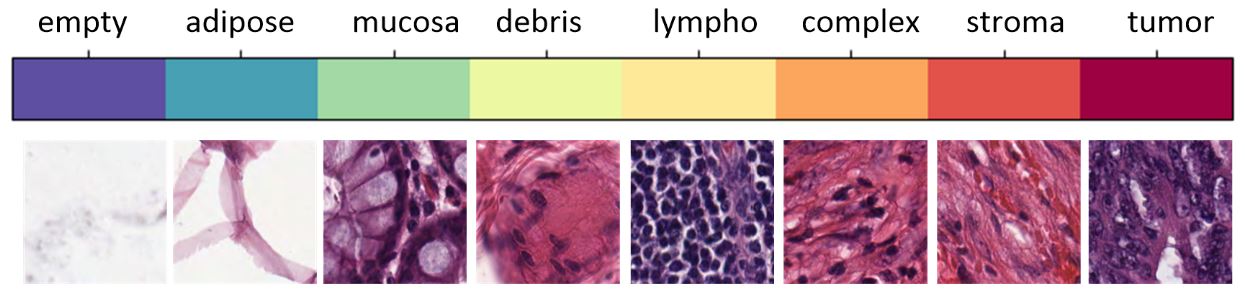}
	\end{subfigure}\\
	\begin{subfigure}{0.24\textwidth}
		\includegraphics[width=\textwidth]{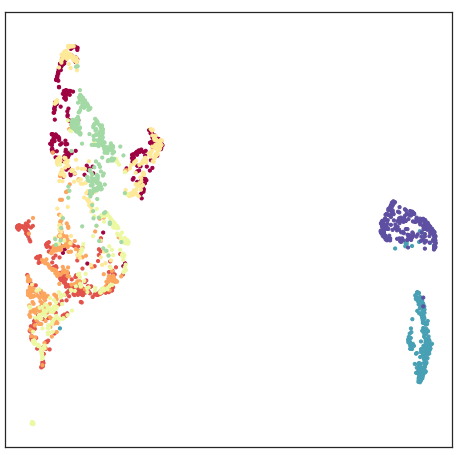}
		\caption{}
	\end{subfigure}
	\begin{subfigure}{0.24\textwidth}
		\includegraphics[width=\textwidth]{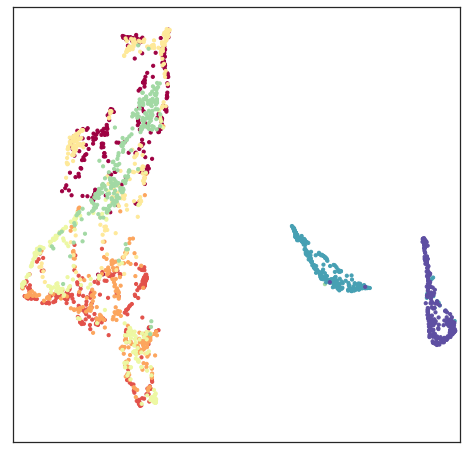}
		\caption{}
	\end{subfigure}
	\begin{subfigure}{0.24\textwidth}
		\includegraphics[width=\textwidth]{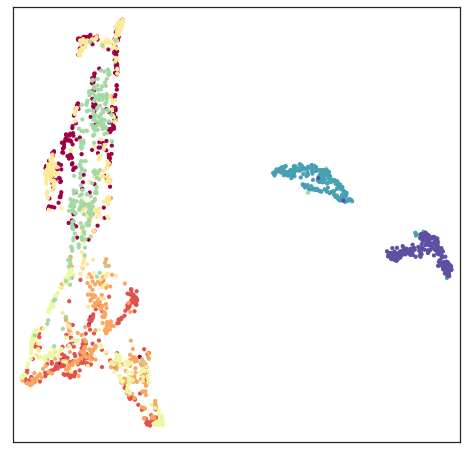}
		\caption{}
	\end{subfigure}
	\begin{subfigure}{0.24\textwidth}
		\includegraphics[width=\textwidth]{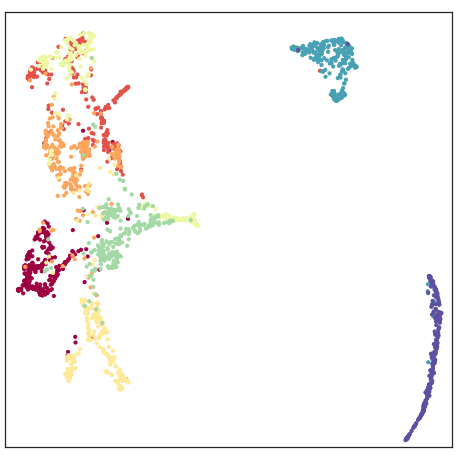}
		\caption{}
	\end{subfigure}
	\caption{$2$-D representation of the $\textrm{D}^{\textrm{T}}$ embedding using models trained by (a) $\textrm{D}^{\textrm{S}}_1$ with the triplet loss, (b) $\textrm{D}^{\textrm{S}}_2$, (c) $\textrm{D}^{\textrm{S}}_3$, and (d) $\textrm{D}^{\textrm{S}}_1$ with the cross-entropy loss. The $2$-D representations are produced by UMAP \cite{mcinnes2018umap}.}
	\label{figure_embeddings_CRC}
\end{figure*}

\section{Related Work}
Koch et al. \cite{koch2015siamese} suggested the Siamese network as a choice for representation learning based on sample-wise comparisons. The Siamese networks generally consist of two similar neural networks that accept two inputs from either the same or different classes. The pair is embedded by identical neural networks first. Then the component-wise difference of the representations is passed into a comparison neural network. By doing so, the Siamese network learns to identify discrepancies between classes. Later, Hoffer and Ailon \cite{hoffer2015deep} proposed triplet networks for representation learning that learned the configuration among anchor, positive, and negative cases (or triplets) simultaneously. In detail, the triplet network learned to put anchor and positive samples closer while pushing the negative samples farther in the latent space. These works fall within a greater field of study called distance metric learning \cite{suarez2018tutorial}. Medela et al. \cite{medela2019few} employed triplet networks on the colorectal cancer slides as a source domain, and utilized their model to extract features and represent data from the healthy and tumorous colon, breast, and lung slides as the target domain. For this purpose, they utilized a VGG16 model \cite{simonyan2014very} as the backbone of their triplet network while they replaced the last fully connected layer with a more compact version. They used labels provided by pathologists in the source domain to create triplets. Furthermore, due to the mismatch between the properties of source and target domain, images were adjusted through rescaling.

%Furthermore, Jean et al. \cite{jean2019tile2vec} employed the triplet network in the context of aerial imaging as an unsupervised representation learning framework for aerial imagery. They utilized a neural network based on ResNet-18 \cite{he2016deep} as the triplet network. Subsequently, they selected a positive sample in close vicinity of the anchor patch to be semantically similar, while the negative example was chosen spatially far enough.

Gildenblat and Klaiman \cite{gildenblat2019self} trained a Siamese network based on a ResNet-50 architecture \cite{he2016deep} by acknowledging adjunct patches as similar and remote tiles as non-similar cases. They both trained and tested their model on the Camelyon16 dataset \cite{bejnordi2017diagnostic}. Eventually, they evaluated their model in a tumor image retrieval task using the Camelyon16 test set. They reported $34\%$ as the ratio of correctly retrieved tumorous patches in comparison with only $26\%$ accuracy using a ResNet-50 \cite{he2016deep} with ImageNet weights. In another work, Teh and Taylor \cite{teh2019learning} investigated the performance of a weakly-supervised framework for representation learning in digital pathology using ResNet-34 \cite{he2016deep}. Thus, they examined the following setups for representation learning with varying target domain size: $(1)$ Training from scratch using target domain dataset, $(2)$ transfer learning with cross-entropy loss based on a network pre-trained on a weakly labeled source dataset, and $(3)$ employing the metric learning approach on the same model, pre-trained with weakly labeled data. Due to the property mismatch between different datasets utilized in their study, the authors had to resize the images. Finally, they reported $92.46\%$ accuracy on using $625$ samples per class, on the CRC dataset as their best outcome \cite{kather2016multi}.

In this study, we aim to address shortcomings such as the definition of the similarity, investigation of the impact of the source and target domain datasets, and the level of supervision in representation learning in digital pathology. The description of the source and target domain datasets, triplet generation algorithm, and representation learning approach are introduced in section \ref{method}. Section \ref{experiments} contains experiments associated with the effects of the source domain, supervision, and target domain size on feature learning. Finally, the discussion and a summary of our findings are described in section \ref{conclusion}.

\section{Methodology}\label{method}

In this work, we used two popular histopathology datasets, namely The Cancer Genome Atlas (TCGA) (available at https://www.cancer.gov/tcga) and colorectal cancer (CRC) \cite{kather2016multi} datasets. The TCGA dataset contains WSIs from 25 different anatomical sites, including 32 different cancer subtypes. We utilized $40\%$ randomly selected WSIs from three organ sites, namely prostate, gastrointestinal, and lung. These organs were chosen as they are among the most commonly diagnosed cancers. These sites had a total of nine cancer subtypes, namely Prostate adenocarcinoma (PRAD), Testicular germ cell tumors (TGCT), Oesophageal carcinoma (ESCA), Stomach adenocarcinoma (STAD), Colonic adenocarcinoma (COAD), Rectal adenocarcinoma (READ), Lung adenocarcinoma (LUAD), Lung squamous cell carcinoma (LUSC), and Mesothelioma (MESO) \cite{cooper2018pancancer}. The CRC dataset \cite{kather2016multi} contains $5,000$ histological images extracted from different tissues present in the colorectal cancer slides. The tissue types are background, adipose tissue, normal mucosal glands, debris, immune cells, complex stroma, simple stroma, and tumor epithelium. Some CRC patches of different tissue types can be found in Fig. \ref{figure_embeddings_CRC}.

\subsection{Triplet Generation}\label{section_triplet_extraction}

\begin{figure}[ht!]
\centering
\includegraphics[width=1\columnwidth,height=1.9in]{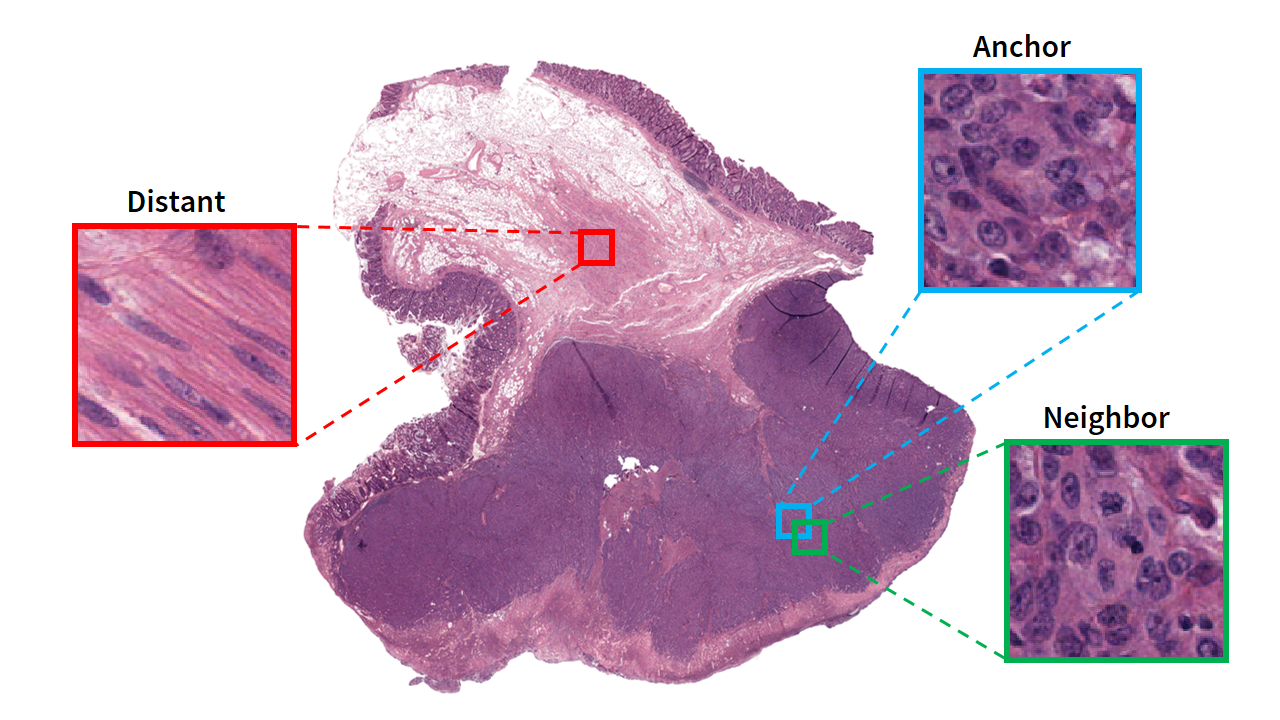}
\caption{An example of the type $1$ triplet generation from a sample WSI (from COAD subtype) from TCGA dataset.}
\label{figure_TCGA_triplet_extraction}
\end{figure}

\begin{figure}[ht!]
	\centering
	\begin{subfigure}{0.3\columnwidth}
		\includegraphics[width=\columnwidth,height=2.7cm]{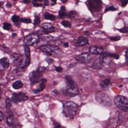}
		\caption{}
	\end{subfigure}
	\begin{subfigure}{0.3\columnwidth}
		\includegraphics[width=\columnwidth,height=2.7cm]{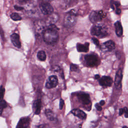}
		\caption{}
	\end{subfigure}
	\begin{subfigure}{0.3\columnwidth}
		\includegraphics[width=\columnwidth,height=2.7cm]{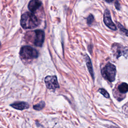}
		\caption{}
	\end{subfigure}\\
	\begin{subfigure}{0.3\columnwidth}
		\includegraphics[width=\columnwidth,height=2.7cm]{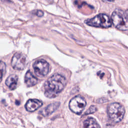}
		\caption{}
	\end{subfigure}
	\begin{subfigure}{0.3\columnwidth}
		\includegraphics[width=\columnwidth,height=2.7cm]{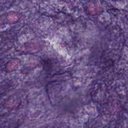}
		\caption{}
	\end{subfigure}
	\begin{subfigure}{0.3\columnwidth}
		\includegraphics[width=\columnwidth,height=2.7cm]{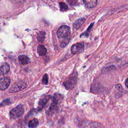}
		\caption{}
	\end{subfigure}\\
	\begin{subfigure}{0.3\columnwidth}
		\includegraphics[width=\columnwidth,height=2.7cm]{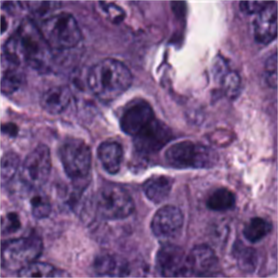}
		\caption{}
	\end{subfigure}
	\begin{subfigure}{0.3\columnwidth}
		\includegraphics[width=\columnwidth,height=2.7cm]{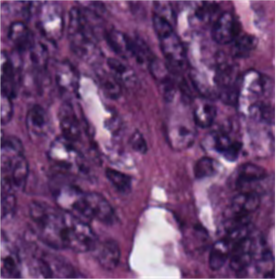}
		\caption{}
	\end{subfigure}
	\begin{subfigure}{0.3\columnwidth}
		\includegraphics[width=\columnwidth,height=2.7cm]{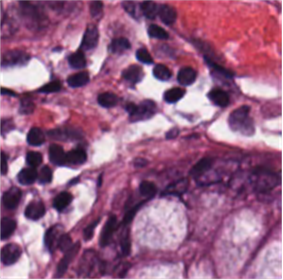}
		\caption{}
	\end{subfigure}
	\caption{The first two rows, sample triplets extracted from TCGA dataset: (a) anchor (STAD), (b) neighbor (STAD), (c) distant type 1 (from the same WSI), (d) distant type 2 (from different STAD WSI), (e) distant type 3 (COAD), and (f) distant type 4 (LUAD). The last row, a sample triplet generated from the CRC dataset \cite{kather2016multi}: (g) anchor, (h) neighbor, and (i) distant.}
	\label{figure_triplets}
\end{figure}

Since we used a triplet network, we defined the similarity concept in a way capable of generating triplets. The triplet was comprised of anchor, neighbor, and distant tiles (patches), in which anchor and neighbor were defined as similar and anchor and distant as dissimilar pairs. Hence, the notion of similarity was abstract enough to avoid limiting the learning performance. Also, the concept of similarity should require minimal supervision to reduce the cost of triplet generation. In this study, similar to \cite{jean2019tile2vec}, we utilized spatial correlation as one of the approaches to define the similarity among patches extracted from WSIs. In other words, we assumed that similar patterns usually emerge in an adjacent neighborhood, while the dissimilar layouts often appear in the spatially remote neighborhood. More specifically, a neighbor patch was selected within a certain range of the anchor's tile center of the same WSI. On the other hand, we used several alternatives for choosing the distant patch. The distant sample was chosen from $(1)$ the same WSI as long as it was spatially remote, $(2)$ another WSI associated with the same cancer subtype, $(3)$ another WSI associated with other subtypes of the same organ, or $(4)$ another WSI associated with another organ. An example of a type $1$ triplet generation is depicted in Fig. \ref{figure_TCGA_triplet_extraction}. For the experiments performed on the CRC dataset \cite{kather2016multi}, we selected the neighbor and distant patches from the same and different tissue types to the anchor patch, respectively. Fig. \ref{figure_triplets} shows sample triplets extracted from the TCGA and the CRC datasets \cite{kather2016multi}.

\subsection{Representation Learning in the Source Domain}

In this paper, we aimed to analyze the effect of the source domain on a few-shot learning framework. 

Therefore, we embedded the data in the source domain using different settings to evaluate its impact on the performance. For embedding, we used the triplet network, with a ResNet-18 backbone.

The triplet loss \cite{schroff2015facenet} was chosen as it evaluates similarity and dissimilarity among samples simultaneously, while the contrastive loss assesses them one by one. We implemented a triplet network \cite{schroff2015facenet} where three ResNet-18 networks \cite{he2016deep} have shared weights. The triplet loss can be defined as:
\begin{equation}
\sum_{i=1}^b \Big[||\hm{f}(\hm{x}_i^a) - \hm{f}(\hm{x}_i^n)||_2^2 - ||\hm{f}(\hm{x}_i^a) - \hm{f}(\hm{x}_i^d)||_2^2 + \alpha\Big]_+,
\end{equation}
where $[z]_+= \max(z, 0)$, $b$ is the batch size, and $\hm{x}_i^a$, $\hm{x}_i^n$, and $\hm{x}_i^d$ are the $i^{\textrm{th}}$ anchor, neighbor, and distant tiles in the batch, respectively. Also, $\hm{f}(\hm{x})$ denotes the embedded output of the network. In the embedding space, the triplet loss pulls the neighbor toward the anchor by minimizing the first term and pushes the distant away from the anchor by maximizing the second term. The term $\alpha$ prevents the network from pushing the distant sample further than the margin value. As a baseline for comparison, we also used a ResNet-18 \cite{he2016deep} trained by a standard cross-entropy loss.

\subsection{Knowledge Transfer to the Target Domain}

After training the triplet network, the target domain data was embedded using the model. Next, a classifier was trained on a portion of the target domain data and was tested against the remaining samples. We used a Support Vector Machine (SVM) as the classifier.

\section{Experiments}
\label{experiments}
For all experiments in this study, we utilized a ResNet-18 \cite{he2016deep} backbone with Adam optimizer (learning rate of $10^{-5}$ and betas of $0.9$ and $0.999$), and a batch size of $32$. Also, all images were of size $128 \times 128$ pixels, extracted from WSIs at $20 \times$ magnification (CRC tiles were cropped to adjust the size). Additionally, the last layer of the model had a dimension of $128$. Also, the margin value, $\alpha$, was set to $0.25$ for all experiments. Furthermore, during the supervised setup, our model was trained using $3,000$ image tiles from the CRC dataset while for the triplet network the total number of $22,528$ triplets was used (both TCGA and CRC dataset). All neural network approaches in this study were implemented using TensorFlow \cite{tensorflow2015-whitepaper}.

\textbf{Triplet Loss vs. Cross-entropy Loss --} First, we split the CRC dataset \cite{kather2016multi} into $60\%$ and $40\%$ portions, denoting $\textrm{D}^{\textrm{S}}_1$ and $\textrm{D}^{\textrm{T}}$, respectively. Accordingly, we trained a ResNet-18 with two different approaches: $(1)$ cross-entropy loss, and $(2)$ triplet loss. To train our model with cross-entropy loss, we attached a softmax layer on top of the neural network. After training both models, the embeddings were extracted from the last layer. The embeddings of the $\textrm{D}^{\textrm{T}}$ with the triplet and the cross-entropy loss are shown in Figs. \ref{figure_embeddings_CRC}-a and \ref{figure_embeddings_CRC}-d, respectively. We applied the Uniform Manifold Approximation and Projection (UMAP) \cite{mcinnes2018umap} to visualize the 128-dimensional representations in $2$D plane. The number of neighborhood parameter was set to $40$ for all $2$-D representations.

\textbf{Source Domain Effect --} The triplets extracted from the CRC training set were sampled in a supervised manner as the labels of tissues were used. However, as we described in Section \ref{section_triplet_extraction}, the triplets of TCGA data were sampled using the spatial and tissue type information in an unsupervised manner. As a result, we trained extra two models on triplets extracted from TCGA. The first one was trained on all three anatomical sites, called $\textrm{D}^{\textrm{S}}_2$, while the second model was only trained on the gastrointestinal data from TCGA called $\textrm{D}^{\textrm{S}}_3$ as the $\textrm{D}^{\textrm{T}}$ was also related to the same anatomical site. Similarly, the $\textrm{D}^{\textrm{T}}$ embeddings encoded by these models are shown in Figs. \ref{figure_embeddings_CRC}-b and \ref{figure_embeddings_CRC}-c.

\begin{table}[ht!]
\caption{Average accuracy and confidence interval of the target domain classification over the different folds of the test dataset.}\label{table_online}
\centering
\scriptsize
\begin{tabular}{cccccccccc}
\hline
Portions of   &  \multicolumn{3}{c}{Triplet}   &&   \multicolumn{1}{c}{Cross-entropy} \\
\cmidrule{2-4} \cmidrule{6-6} \vspace{1mm}
Data (\%)    &  \textrm{$D_{1}^S$} &  \textrm{$D_{2}^S$ }  &  \textrm{$D_{3}^S$}  &&   \textrm{$D_{1}^S$ }  \\
\hline
\textbf{5}   & 83.00$\pm$2.18 & 77.00$\pm$2.52 & 75.00$\pm$2.98 && 91.00$\pm$1.65 \\
\textbf{10}  & 88.50$\pm$0.99 & 88.50$\pm$0.77 & 88.50$\pm$0.77 && 87.00$\pm$1.09 \\
\textbf{25}  & 92.80$\pm$0.41 & 93.20$\pm$0.24 & 94.20$\pm$0.29 && 87.00$\pm$0.35 \\
\textbf{50}  & 94.90$\pm$0.11 & 93.80$\pm$0.19 & 94.60$\pm$0.14 && 89.50$\pm$0.14 \\
\textbf{100} & 95.90$\pm$0.03 & 95.75$\pm$0.07 & 95.95$\pm$0.09 && 91.90$\pm$0.09 \\ 
\hline
\end{tabular}
\end{table}

\textbf{Target Domain Size Effect --} Finally, we split the $\textrm{D}^{\textrm{T}}$ into chunks of $5\%$, $10\%$, $25\%$, and $50\%$ using stratified sampling, besides the full set. Then we trained an SVM classifier utilizing 10-fold cross-validation on the latter subsets. For all experiments, all SVM classifiers were fine-tuned by changing the kernel (linear, RBF, sigmoid, and polynomial), and the parameters C (between $0.001$, and $1000$) and gamma (between $0.001$ and $1000$, scale and auto) to achieve the best performance. All results are reported in Table \ref{table_online}.

\section{Summary and Conclusions}
\label{conclusion}

As it is depicted in Fig. \ref{figure_embeddings_CRC}, empty and adipose tiles' representation were closely located to each other and far from other tissue textures in all models. Also, the mucosa was almost well-clustered in all embeddings because of its unique pattern. Moreover, the corresponding representations of classes with visual similarities blended in $2$-D space when the triplet loss was employed. According to Table \ref{table_online}, however, all models trained by the triplet loss scored higher in terms of mean accuracy in comparison to the model trained with the label information using cross-entropy loss, excluding $5\%$ subset. The confidence intervals shrunk as the size of the target data increased. Furthermore, the confidence interval was relatively larger when the source domain was different from the target domain. More importantly, the highest accuracy, $95.95\%$, was achieved using the model which was trained on the gastrointestinal subset of the TCGA. This outcome may suggest that extensive training on a similar dataset using weakly supervised and unsupervised methods can improve the performance of the model and generalization of the solution.

In this work, We investigated the effect of the supervision and the source domain in embedding and few-shot learning for histopathology images. Accordingly, we have shown that a network can learn a meaningful representation from histopathology images using a massive amount of the weakly labeled data currently available online. This was achieved by acknowledging the spatial correlation, anatomical information, and slide level diagnosis.
\balance

\bibliographystyle{IEEEtran}
\bibliography{references}
\end{document}